# Réalisation d'un système de reconnaissance automatique de la parole arabe base sur CMU Sphinx

**Ali Sadiqui, Noureddine Chenfour**
Faculté des Sciences Dhar El Mehraz,
Université Sidi Mohamed Ben Abdellah de Fès
B.P. 1796 – Fès – Morocco

**ABSTRACT.** This paper presents the continuation of the work completed by Satori and all. [SCH07] by the realization of an automatic speech recognition system (ASR) for Arabic language based SPHINX 4 system. The previous work was limited to the recognition of the first ten digits, whereas the present work is a remarkable projection consisting in continuous Arabic speech recognition with a rate of recognition of surroundings 96%.

## Introduction

Le développement d'un système de reconnaissance automatique de la parole (ASR) arabe a récemment suscité l'intérêt d'un certain nombre de chercheurs. Plusieurs tentatives de construction d'un ASR arabe (AASR) ont été notées, et elles ont abouti à des résultats encourageants. [AB02, AEA09, P+88, TAM07] Cependant la plupart de ces tentatives s'est limité à un vocabulaire réduit.

Dans ce document, nous présentons les résultats obtenus pour la réalisation AASR dépendant du locuteur à grand vocabulaire et basé sur la langue arabe en utilisant le logiciel CMU Sphinx 4.

Nous avons abouti à développer un ASR mono locuteur (speaker dependant system) avec un taux de reconnaissance de 96% (WER = 0.4).

Dans la première section de cet article, nous allons commencer à présenter les Modèles de Markov Cachés pour expliquer les deux composants d'un ASR à savoir le modèle acoustique et le modèle de langage. Dans la deuxième section nous présenterons le logiciel SPHINX, et pour finir nous présenterons le système réalisé, ainsi qu'une analyse des résultats avant de conclure.

**1. Les Modèles de Markov Cachés**

Dans l'état actuel des connaissances, les modèles des Markov cachés (HMM) sont les outils de modélisation les plus utilisés en matière de reconnaissance de la parole. Leur application débouche sur d'excellents résultats quand ils sont bien appliqués [Rab89].

Généralement, Les ASR sont composé de deux éléments essentiels : le modèle acoustique et le modèle du langage.

Nous pouvons définir les deux modèles comme suit:
- Le modèle acoustique regroupe l'ensemble des informations concernant la représentation phonétique, la variabilité de l'environnement du locuteur, du sexe du locuteur etc. [HAH01].
- Le modèle de langage a pour objectif de répondre aux contraintes du langage naturel, surtout lorsqu'il s'agit de mots qui ont la même prononciation, et améliorant ainsi leur décodage [Mar90].

Théoriquement nous pouvons expliquer ces deux modèles de la façon suivante:

Dans une approche stochastique, si nous considérons un signal acoustique S, le principe de la reconnaissance peut être expliqué comme le calcul de la probabilité P(W|S) qu'une suite de mots (ou phrase) W correspond au signal acoustique S, et de déterminer la suite de mots qui peut maximiser cette probabilité.

En utilisant la formule de Bayes, P(W|S) peut s'écrire :

$$P(W|S) = P(W).P(S|W)/P(S)$$

avec:
* P(W) est la probabilité a priori de la suite de mots W.
* P(S|W) est la probabilité du signal acoustique S, étant donné la suite de mots W
* P(S) est la probabilité du signal acoustique (que nous supposons, par simplification, ne pas dépendre de W).

P(S|W) est nommé Modèle Acoustique, et P(W) est nommé Modèle de Langage.

Ces deux modèles peuvent être représentés comme un Modèle Markovien.

Un modèle HMM est défini par l'ensemble de données suivantes:

Une matrice A qui indique les probabilités de transition d'un état qi vers un autre état (ou vers lui-même), soit p(qj|qi)= $a_{ij}$.

Une matrice B qui indique les probabilités d'émission des observations dans chaque état. Si nous considérons le cas de la parole continue, cette probabilité est de type multigaussienne, définie par les vecteurs moyens, les matrices de covariance et des poids associés à chaque gaussienne.

Une matrice Π donne la distribution de départ des états, c'est-à-dire pour chaque état la probabilité d'être atteint à partir de l'état initial $q_i$. Cet état est particulier puisqu'il ne peut émettre d'observations.

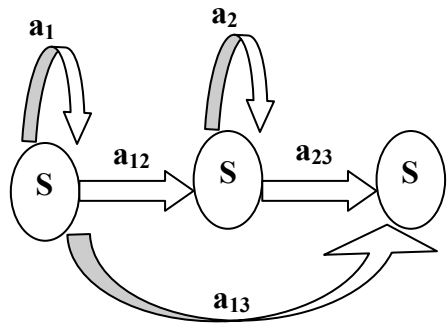

**Figure 1: Modèle HMM dit gauche-droit d'ordre 1 à 3 états**

Cette structure rend possible la représentation des changements dus à la prononciation. En effet pour l'articulation lente, il y a répétition d'états. Il sera représenté dans un modèle HMM par une transition d'état sur lui même (la transition $a_{ii}$ sur la figure précédente). Alors que pour les articulations rapides, le modèle admet le saut à l'état suivant (transition a13).

Parmi les produits réussis dans le domaine de la reconnaissance de la parole, figure l'outil CMU Sphinx 4. Ses points forts résident, entre autres, dans son code ouvert qui offre une grande souplesse d'utilisation, et son appui sur le modèle HMM.

## 2. L'Outil CMU Sphinx

Le système CMU Sphinx est une série d'outils servant à construire des applications de reconnaissance vocale. Au cours des dernières années, elle comprend également des ressources connexes, telles que des modèles acoustiques et des modèles de langages [C+07b, Cmu09].

Le CMU Sphinx comprend entre autres les outils suivants**:**
1. Sphinx 2: est un système de reconnaissance de la parole à grande vitesse. Il est habituellement employé dans des systèmes de dialogue et des systèmes d'étude de prononciation.
2. Sphinx 3: est un système de reconnaissance de la parole légèrement plus lent mais plus précis.
3. Sphinx 4: Une réécriture complète du Sphinx en Java. Il offre à la fois la précision et la rapidité.
4. Sphinxtrain: Une suite d'outils qui permet de créer le modèle acoustique
5. CMU-Cambridge Language Modeling Toolkit: Une suite d'outils qui permet de créer le modèle de langage.

La figure 2 montre l'évolution de l'outil CMU Sphinx4.

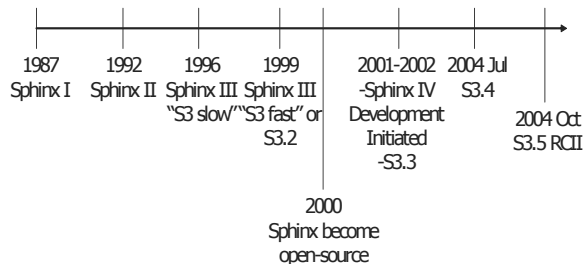

**Figure 2: L'évolution des versions de SPHINX [C+07a]**

La disponibilité de code source Sphinx4 rend possible sa flexibilité et encourage la recherche dans les universités et les laboratoires spécialisés (HP, Sun…)**.**

Le logiciel Sphinx4 utilise à la fois le modèle acoustique et le modèle de langage pour décoder un signal acoustique.

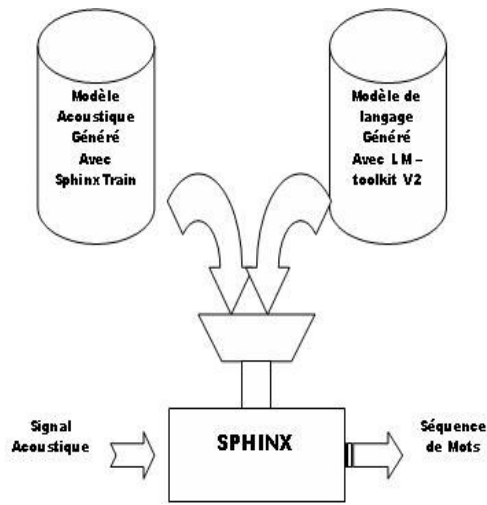

**Figure 3 :** **Schéma simplifiant le traitement en utilisant le logiciel Sphinx 4**

## 3. Description du système

### 3.1. Corpus

Notre corpus est constitué de 1443 fichiers décomposés de la façon suivante:

**Tableau 1. La décomposition de corpus utilisée**

|  | Série #1 | Série #2 |
|---|---|---|
| Nombre de fichier | 768 | 675 |
| Locuteur | Voix de femme | Voix de la même femme |
| Moyenne de la durée d'un fichier | 21.69 sec | 5.83 sec |
| Type d'utilisation | Traitement | Test |
| Durée Total | 4h 37 min | 1h 4 min |

La durée totale du corpus est donc de 5h41min dont ≈ 20% a été utilisé pour le test des performances et le reste pour la création de modèle acoustique.

La totalité du corpus est composé de voix d'une seule femme.

La transcription a été faite de sorte qu'elle s'adapte à la prononciation du locuteur même s'elle contient des erreurs de grammaire. A noter aussi

31

que nous avons utilisé des phonèmes inexistants dans l'arabe traditionnel, comme les phonèmes v, p et g.

## 3.2. Création du Modèle acoustique

Le but de l'analyse acoustique consiste à représenter le signal de parole sous une forme qui est plus adaptée pour la reconnaissance. Le plus souvent les représentations suivantes son utilisées: MFCC (Mel Frequency Cepstral Coeficients en anglais), LPCC (Linear Predictive Cepstral Coefcients) ou PLP (Perceptual Linear Predictive analysis). Dans ce travail nous avons utilisé des paramètres acoustiques de type MFCC (12MFCC+E) et leurs dérivées première et seconde. Ces vecteurs sont normalisés par rapport à la moyenne et la variance sur une phrase. La normalisation par rapport à la variance permet de diminuer la variabilité par rapport au locuteur. L'extraction des paramètres est réalisée avec l'outil Wave2feat de SPHINX.

Pour la modélisation acoustique, nous avons utilisé comme unité de base des phones en contexte. Chaque phone en contexte est modélisé par une chaîne de Markov cachée à 3 états avec des densités d'observation multi-gaussiennes.

La procédure pour créer le modèle acoustique consiste à regrouper un ensemble de données d'entrées, appelé la base de données de traitement, et de les traiter avec l'outil SphinxTrain.

Les données d'entrées sont composées, entre autre :
- d'un ensemble de fichiers acoustiques (corpus).
- d'un fichier de transcription qui contient l'ensemble de mots prononcés pour chaque enregistrement (fichier acoustique).
- d'un fichier dictionnaire qui définie la décomposition phonétique pour chaque mot.
- d'un fichier qui définie la liste des phonèmes utilisées.

Sphinxtrain est l'outil destiné à créer le modèle acoustique. Il est composé d'un ensemble de script PERL en outre des fichiers de configuration.

Nous avons procédé à une notation latine pour représenter les mots ainsi que les phonèmes utilisés.

Notre fichier des phonèmes contient 37 phonèmes.

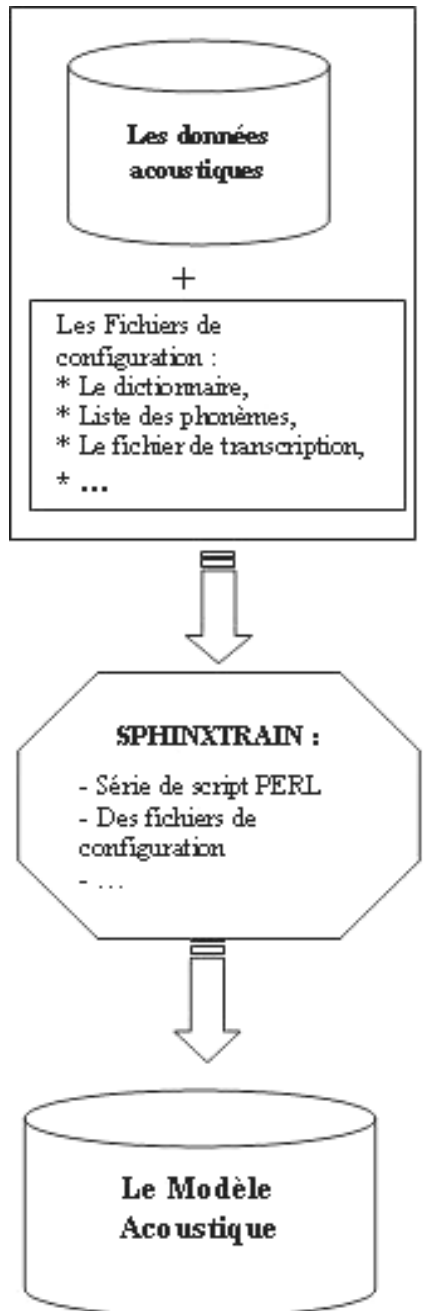

**Figure 4 : Schéma représentant la création d'un modèle acoustique avec Sphinxtrain**

Le tableau suivant représente la notation des phonèmes utilisée :

**Tableau 2. Notation des phonèmes**

| Phoneme | Notation | Phoneme | Notation |
|---|---|---|---|
| ء | @ | ق | q |
| ب | b | ك | k |
| ت | t | ل | l |
| ث | t_h | م | m |
| ج | j | ن | n |
| ح | ~h | هـ | h |
| خ | x | و | w |
| د | d | ي | y |
| ذ | z2 | الجيم المصرية | g |
| ر | r | اسبانيا >-p | p |
| ز | z | نوفبر >-v | v |
| س | s | الفتحة -> بَ | a |
| ش | s_h | المد-> با | a2 |
| ص | s2 | الضمة -> بُ | u |
| ض | d2 | المد -> بو | u2 |
| ط | t2 | الكسرة -> بِ | i |
| ظ | ~z | المد -> بي | i2 |
| ع | ~@ | الشدة | اعادة الحرف |
| غ | g_h | التنوين | a n /u n / i n |
| ف | f | | |

**Tableau 3. Notation de mots et leur décomposition phonétique**

| Mot arabe | Notation latine | Décomposition phonétique |
|---|---|---|
| وسيمون | wasiymoun | W A S I2 M U2 N |
| غالبا | ghaaliban | G_H A2 L I B A N |
| سجود | sojoud | S U J U2 D |
| أحرز | ?ux\riza | @ U ~H R I Z A |
| رقبة | raqaba | R A Q A B A |
| قوة | quwwa | Q U W W A |
| نقود | noqoud | N U Q U2 D |
| متبادل | mutaba:dal | M U T A B A2 D A L |
| مسلم | muslim | M U S L I M |
| عهد | ?`ahd | ~@ A H D |
| بضبط | bid`d`abt` | B I D2 D2 A B T2 |

Notre dictionnaire lors de traitement est constitué de 12822 mots.

Exemple de notation adoptée dans notre dictionnaire sont présentée dans le Tableau 3.

### 3.3. Création du modèle de langage

Le modèle utilisé dans cette étude à été crée avec les outils « The CMU-Cambridge Statistical Language Modeling Toolkit v2 ».

Le corpus utilisé pour ledit modèle contient un dictionnaire de 15786 mots.

Nous nous sommes limités à créer des modèles à bigramme. Le nombre d'unigramme = 15788 et de bigramme = 27387.

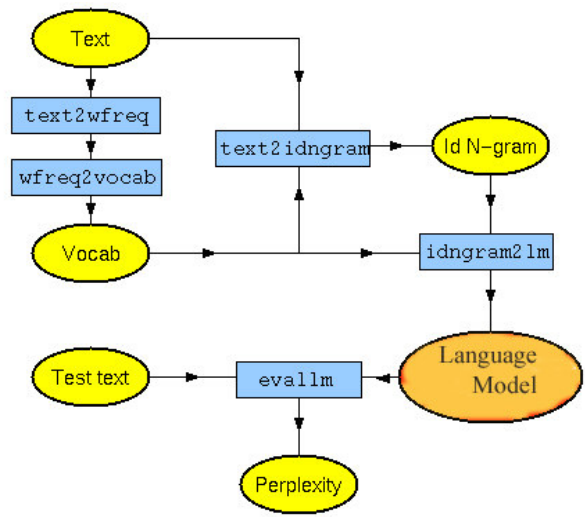

**Figure 5: Schéma représentant les étapes pour la création d'un modèle de langage avec "CMU-Cambridge Language Modeling Toolkit v2" [Spe09]**

### 3.4. Résultats

L'évaluation du système réalisé est faite à partir d'un corpus composé de 675 fichiers. Les résultats en termes de temps et en termes de mots sont présentés dans les tableaux 4 et 5.

35

**Tableau 4. Résultats en termes de temps**

| Le temps total des fichiers de test | Le temps total des fichiers totalement décodés | Taux de reconnaissance |
|---|---|---|
| 1h 4 min | 52 min | 82% |

**Tableau 5. Résultats en termes de mots**

| Total des mots testés | Total des mots décodés | Taux de reconnaissance |
|---|---|---|
| 5354 | 5167 | 96 % |

Remarque: À noter que les des fichiers mal décodé contiennent des mots correctement décodés.

Le test a abouti à des résultats très satisfaisants et la performance du système réalisé est démontrée. Les mots correctement identifiés représentent 96%. Le nombre de cas de substitution de mot est de 134, et les erreurs d'insertion sont de 48. Davantage d'analyse indique que la majorité des erreurs de substitution de mot est due à de légères différences.

**Table 6. Les performances de moteur de reconnaissance Sphinx sous différent type de corpus [Cmu09]**

| La taille de vocabulaire | Test | % WER |
|---|---|---|
| 11 | TIDIGITS | 0.661 |
| 79 | AN4 | 1.300 |
| 1,000 | RM1 | 2.746 |
| 5,000 | WSJ5K | 7.323 |
| 60,000 | HUB4 | 18.845 |

Le tableau ci-dessus représente le nombre de fois que les phonèmes ont été testés et le % de leur reconnaissance.

**Tableau 7. Tableau représentant le nombre de fois que les phonèmes ont été testés et le % de leur reconnaissance**

| Phonème | # testé | # décodé | % décodage |
|---|---|---|---|
| @ | 1225 | 1182 | 96,49 |
| A | 6190 | 6051 | 97,75 |
| a2 | 2134 | 2089 | 97,89 |
| B | 953 | 924 | 96,96 |
| D | 846 | 826 | 97,64 |
| d2 | 200 | 194 | 97,00 |
| F | 696 | 680 | 97,70 |
| G | 15 | 14 | 93,33 |
| g_h | 105 | 104 | 99,05 |
| H | 757 | 737 | 97,36 |
| I | 3733 | 3653 | 97,86 |
| i2 | 852 | 832 | 97,65 |
| J | 364 | 362 | 99,45 |
| K | 549 | 538 | 98,00 |
| L | 2605 | 2528 | 97,04 |
| M | 1558 | 1522 | 97,69 |
| N | 1827 | 1782 | 97,54 |
| P | 42 | 39 | 92,86 |
| Q | 544 | 538 | 98,90 |
| R | 1525 | 1491 | 97,77 |
| S | 753 | 736 | 97,74 |
| s2 | 298 | 284 | 95,30 |
| s_h | 364 | 355 | 97,53 |
| T | 1820 | 1795 | 98,63 |
| t2 | 253 | 244 | 96,44 |
| t_h | 240 | 236 | 98,33 |
| U | 1873 | 1838 | 98,13 |
| u2 | 524 | 514 | 98,09 |
| V | 22 | 21 | 95,45 |
| W | 812 | 792 | 97,54 |
| X | 216 | 208 | 96,30 |
| Y | 1527 | 1480 | 96,92 |
| Z | 221 | 221 | 100,00 |
| z2 | 69 | 65 | 94,20 |
| ~@ | 788 | 769 | 97,59 |
| ~h | 530 | 518 | 97,74 |
| ~z | 130 | 123 | 94,62 |

**Conclusion**

Le système que nous avons réalisé a montré des résultats très satisfaisants avec un taux de reconnaissance assez élevé. Nous considérons cependant, qu'il s'agit d'un travail préliminaire qui nous permettra par la suite l'accomplissement de notre objectif de base ; un système de reconnaissance indépendant du locuteur pour la langue arabe. Dans cette perspective, nous projetons de préparer un corpus plus important dont l'étude a déjà été effectuée.

**Bibliographie**